\documentclass{article}
\usepackage{spconf,amsmath,epsfig}
\usepackage[T1]{fontenc}
\usepackage[utf8]{inputenc}
\usepackage{tabularx}
\usepackage{booktabs}
\usepackage{multirow}
\usepackage[capitalize]{cleveref}
\usepackage{subcaption}

\newcolumntype{Y}{>{\centering\arraybackslash}X}

\title{Generative adversarial networks for realistic synthesis of hyperspectral samples}
\name{Nicolas Audebert \textsuperscript{1, 2},
	  Bertrand Le Saux \textsuperscript{1},
      Sébastien Lefèvre \textsuperscript{2}
\thanks{N. Audebert's work is funded by the Total-ONERA project NAOMI.}}
\address{\textsuperscript{1} ONERA, \textit{The French Aerospace Lab}, F-91761 Palaiseau, France\\
		 \textsuperscript{2} Univ. Bretagne-Sud, UMR 6074, IRISA, F-56000 Vannes, France}
\begin{document}
%
\maketitle
\begin{abstract}
This work addresses the scarcity of annotated hyperspectral data required to train deep neural networks. Especially, we investigate generative adversarial networks and their application to the synthesis of consistent labeled spectra. By training such networks on public datasets, we show that these models are not only able to capture the underlying distribution, but also to generate genuine-looking and physically plausible spectra. Moreover, we experimentally validate that the synthetic samples can be used as an effective data augmentation strategy. We validate our approach on several public hyperspectral datasets using a variety of deep classifiers.
\end{abstract}
\section{Introduction}
\label{sec:intro}


Data augmentation consists of introducing unobserved samples into the optimization process of a statistical model~\cite{dyk_art_2012}. Since the reintroduction of Convolutional Neural Networks (CNN) for image classification~\cite{krizhevsky_imagenet_2012}, this practice has been critical to avoid overfitting of deep networks.
Therefore, as annotated hyperspectral data is scarce, overfitting is an even more common pitfall compared to multimedia image processing. Although recent efforts have been made to use CNN for hyperspectral image classification~\cite{chen_deep_2016,makantasis_deep_2015,slavkovikj_hyperspectral_2015,lee_contextual_2016}, successes are limited to small datasets that do not leverage the generalization capacity of deep networks.

To this end, recent works have started to investigate data augmentation as a way to artificially enlarge the quantity of annotated samples. For example,~\cite{windrim_hyperspectral_2016} suggested a model of relighting to simulate the same hyperspectral pixel under different illuminations. With a more data-driven approach,~\cite{acquarelli_convolutional_2017} introduced a label propagation strategy to incorporate observed but unlabeled samples to the training set. However, these methods require either unlabeled samples or physics-related assumptions and modeling. On the other hand,~\cite{gemp_inverting_2017} introduced generative models in hyperspectral image processing by using variational autoencoders to find the endmembers composition of spectral mixtures.

In this work, we introduce a way to artificially synthesize new annotated hyperspectral samples using a purely data-driven approach based on generative adversarial networks~\cite{goodfellow_generative_2014}. More specifically, we use a GAN to approximate the distribution of the observed hyperspectral samples and to generate new plausible samples that can be used to train deep networks. Our method can exploit both labeled and unlabeled samples and is validated on several datasets covering both aerial and satellite sensors over rural and urban areas.

\section{Generative models}
\label{sec:generative}

\begin{figure}
	\includegraphics[width=0.49\textwidth]{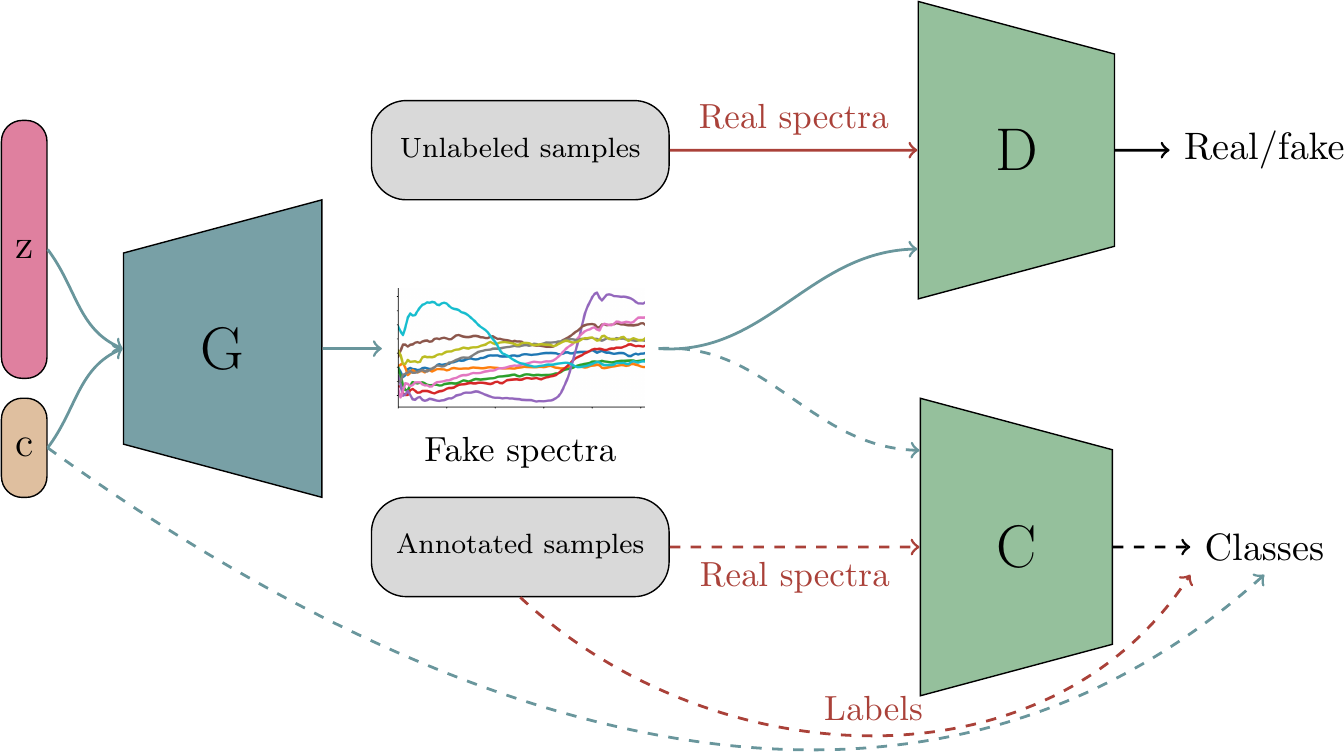}
    \caption{The generative adversarial network used for hyperspectral sample synthesis. Red arrows denote the training of the classifier and discriminator, while the blue arrows denote the training of the generator. Dashed lines denote connections used in the supervised setting.}
    \label{fig:gan}
    \vspace*{-1em}
\end{figure}

The Generative Adversarial Network framework has been introduced in~\cite{goodfellow_generative_2014}. It uses deep neural networks to approximate an unknown data distribution based on its observations. The idea is to generate new samples of a given distribution by training a generator to map random noise from the latent space to the distribution. However, the target distribution is observed only on some data points and we wish to use the generator to create new data points that also still belong to the underlying distribution. To this end, the generator is trained to approximate the distribution using an adversarial objective function. This function is obtained by introducing another network called the discriminator -- or critic. The discriminator learns to infer whether a given sample belongs to the true or the fake distribution, i.e. if the sample belongs to the training set or was created by the generator. The discriminator is trained for a few steps, and then the generator is optimized to fool the critic, i.e. to generate samples that are indistinguishable to the discriminator.

Several flavors of GAN have been introduced that use various objective functions. In this work, we use a generator $G$ and a discriminator $D$ in the Wasserstein GAN~\cite{arjovsky_wasserstein_2017} fashion, trained with the gradient penalty from~\cite{gulrajani_improved_2017}. However, this GAN setup alone only makes it possible to infer a global distribution. In our case, we wish to condition the output of the generator w.r.t. the hyperspectral classes. More specifically, we want our generator to take as an input a random noise and class label, so that it learns to generate a sample beloging to the specified class. This is called conditioning a GAN. To do so, we introduce an additional classifier network $C$. This classifier adds a conditional penalty on the generated distribution by enforcing that the generated spectra are classified in the same class as the conditional label $G$ that was given. The whole framework is illustrated in~\cref{fig:gan}. While $G$ and $D$ can be trained without label knowledge (i.e. unsupervised training), $C$ needs annotated samples to learn.

\section{Experimental setup}
\label{sec:experimental_setup}

We train our GAN on four datasets: Pavia University and Pavia Center (urban aerial scenes at 1.3m/px and 103 bands), Indian Pines  (agricultural scene at 20m/px and 224 bands) and Botswana datasets (swamps, acquired by the Hyperion sensor at 30m/px with 242 bands). We use atmospheric correction when available and we normalize the reflectance between $[0, 1]$. As we try to approximate individual hyperspectral pixels with no spatial context, we use 4-layers deep fully connected networks with 512 neurons for $G$, $D$ and $C$ using the leaky ReLU non-linearity~\cite{maas_rectifier_2013}. $G$ is followed by a sigmoid activation and outputs a vector which length equals the expected number of bands, while $C$ has as many outputs as classes and $D$ only has one output.

Optimization is done for all three networks using the RMSprop stochastic gradient descent policy. The GAN is trained for 100,000 iterations, with $C$ and $D$ being trained twice per iteration before optimizing $G$'s weights.

\section{Spectra analysis}
\label{sec:analysis}

In this section, we aim to investigate the physical plausibility of the synthetic spectra. Especially, we compare the real and fake distributions under several criteria. To this end, we train two GANs on random samples from the Pavia University and Indian Pines datasets.

\begin{figure}
\begin{subfigure}{0.235\textwidth}
\includegraphics[width=\textwidth]{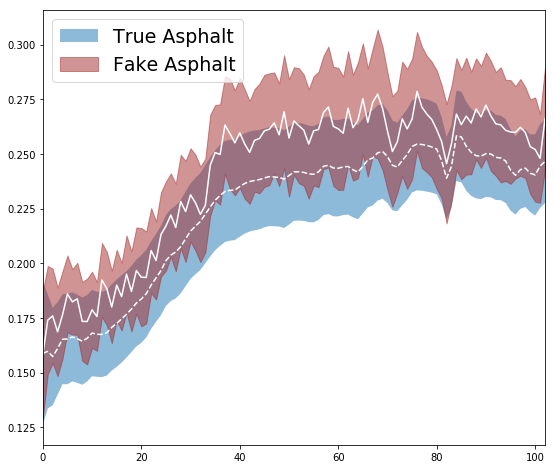}
\end{subfigure}
\begin{subfigure}{0.235\textwidth}
\includegraphics[width=\textwidth]{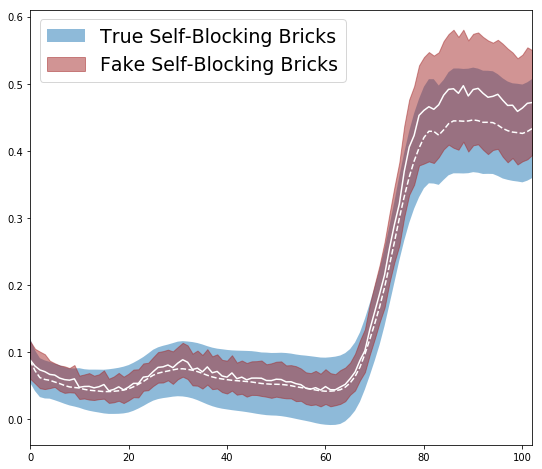}
\end{subfigure}
\caption{Mean spectrum and standard deviation per class for two classes of the Pavia Center dataset. Fake spectra look noisier as they overfit on local spectral properties.}
\label{fig:mean_spectra}
\end{figure}

We can visually assess the quality of the generated spectra by comparing their statistical moments, e.g. plotting the mean spectra and their standard deviation (\cref{fig:mean_spectra}). As can be seen, the spectral shapes are accurately learned by the GAN. However, we can immediately identify two potential shortcomings. First, the fake mean spectra appear noisier than the true spectra, which means that the GAN overfitted on some specific features that are common to only a subset of the real spectra. Second, the fake standard deviation is lower than the real one, which means that fake spectra are less diverse than the real ones. Both of those signs point to a form of overfitting called mode collapse~\cite{salimans_improved_2016}.

\begin{figure}[t]
\begin{subfigure}{0.49\textwidth}
\includegraphics[width=\textwidth]{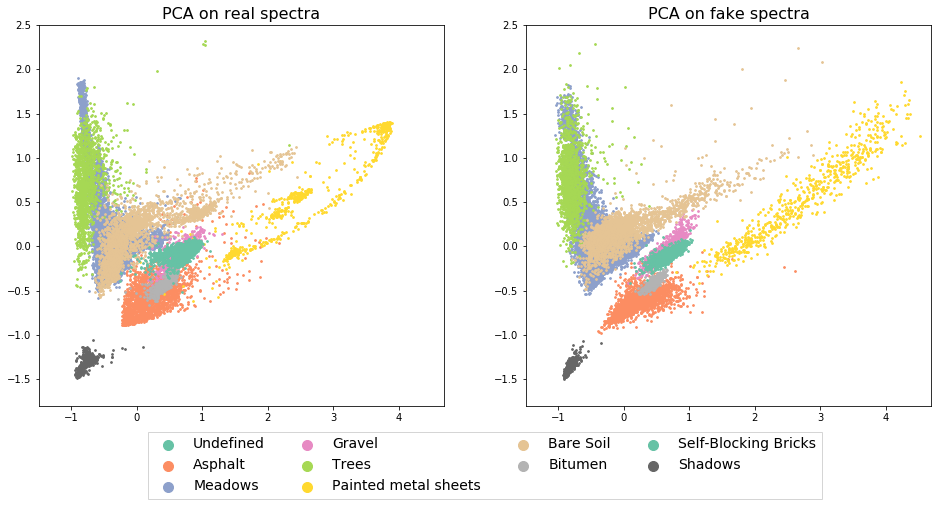}
\caption{Pavia University}
\end{subfigure}
\begin{subfigure}{0.49\textwidth}
\includegraphics[width=\textwidth]{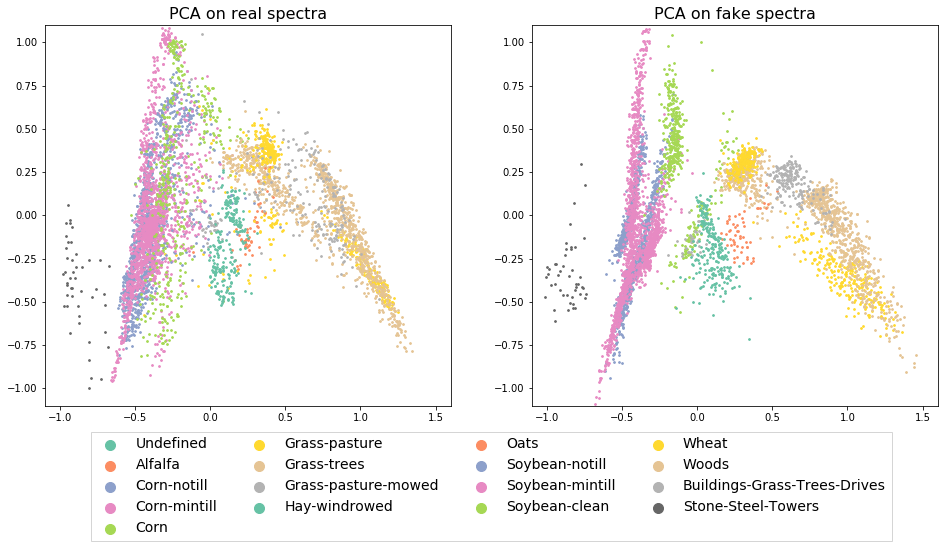}
\caption{Indian Pines}
\end{subfigure}
\caption{PCA applied on real and fake spectra. Real spectra have been randomly sampled from the annotated part of the image. The two sets have the same number of samples.}
\vspace*{-1em}
\label{fig:pca}
\end{figure}

To learn more about how this overfitting actually impacts the distribution of the fake samples in the spectral space, we apply Principal Component Analysis (PCA) to map the spectra into a 2D space (~\cref{fig:pca}). The clusters formed by the different classes are reproduced truthfully by the synthesized samples. However, there are slight deformations that show that the GAN failed to capture some specificities of each class.

\begin{table}[t]
	\begin{tabularx}{0.49\textwidth}{c Y Y Y Y}
        Split & \multicolumn{2}{c}{Random (uniform)} & \multicolumn{2}{c}{Disjoint}\\
        \toprule
		\textbf{Train $\backslash$ Test} & Real & Fake & Real & Fake\\
        \midrule
        Real & 89.5 & 98.3 & 87.2 & 98.8\\
        Fake & 87.8 & 99.2 & 79.4 & 99.9\\
        \bottomrule
	\end{tabularx}
    \caption{Accuracies of a linear SVM on real and fake samples from the Pavia University dataset.}
    \label{table:svm_separation}
\end{table}

We can form an intuition on how the fake distribution respects the class boundaries of the real spectra by training a linear Support Vector Machine (SVM) on the latter and applying it on the former. The SVM will learn the best separating hyperplanes from the true distribution. Hopefully, these hyperplanes should separate the synthesized spectra with the same accuracy. If the accuracy is significantly lower, then the GAN learned unrealistic samples; if it is significantly higher, then the GAN learned samples too similar to the center of each class cluster, i.e. suffered from mode collapse. Results are presented in~\cref{table:svm_separation}. We consider two train/test splits: either 3\% of randomly selected annotated samples or two disjoint halves of the image, i.e. spatially disjoint sets of 50\% of the pixels. In the unsupervised setting, we also use the unlabeled samples. As expected, it is easier for the SVM to separate the fake data than the real samples. However, training on fake samples only still reach encouraging accuracies, only between 2\% and 8\% under the reference real/real setting. This means that although the synthesized spectra are concentrated around the main mode of each class, they still are representative of their class.

\begin{figure}[!t]
\begin{subfigure}{0.49\textwidth}
\includegraphics[width=0.95\textwidth]{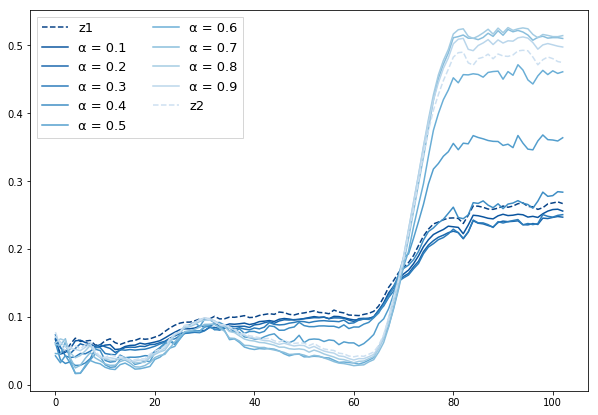}
\caption{Interpolation between two samples of the ``bare soil'' class.}
\end{subfigure}

\begin{subfigure}{0.49\textwidth}
\includegraphics[width=0.95\textwidth]{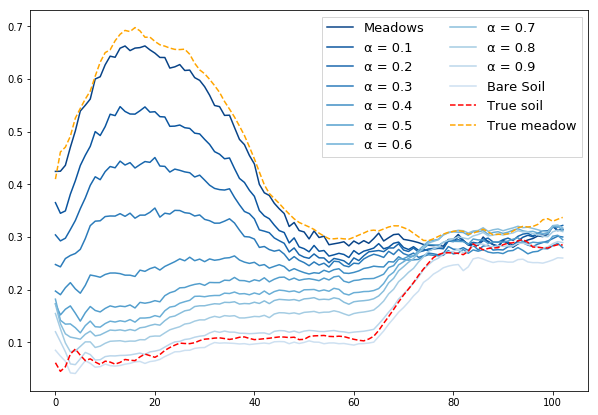}
\caption{Interpolation with a fixed noise vector between the ``meadows' and ``bare soil'' classes.}
\end{subfigure}
\caption{Interpolations in the latent noise space allow the generator to synthesize a continuous variety of samples in the spectral domain. The GAN was trained on the Pavia University dataset. $\alpha$ controls the interpolation.}
\label{fig:interpolation}
\vspace*{-1em}
\end{figure}

Finally, as GANs map a latent noise space to the signal space, it is possible to explore the spectral manifold by interpolating between two noise vectors. Within a fixed class, it allows to generate spectra between two arbitrary points of the latent space. However, it is also possible to interpolate between two classes to generate intermediate spectra that do not necessarily belong to one specific class. This is illustrated in~\cref{fig:interpolation}. There is a continuous progression between the origin and target vectors, which is especially interesting in the inter-class interpolation. The generator learns to perform realistic mixing of several materials, which is the reverse of the unmixing task. Dictionary learning, nearest-neighbors or reversibility approaches such as~\cite{gemp_inverting_2017} could be used to retrieve the material mixing if an exhaustive panel of synthetic mixes has been generated.

\begin{table*}[!t]
	\begin{tabularx}{\textwidth}{c c Y Y Y Y Y Y Y Y}
	\toprule
    \multicolumn{2}{c}{Dataset} & \multicolumn{2}{c}{PaviaU} & \multicolumn{2}{c}{PaviaC} & \multicolumn{2}{c}{Botswana} & \multicolumn{2}{c}{Indian Pines}\\
    Classifier & Augmentation & 3\% (r) & 50\% (s) & 3\% (r) & 50\% (s) & 3\% (r) & 50\% (s) & 3\% (r) & 50\% (s)\\
    \midrule
    \multirow{4}{*}{NN 1D} & $\emptyset$ & 92.72 & 86.22 & 98.93 & 96.26 & 86.90 & 84.87 & 79.44 & 74.00\\
    & GAN & 92.95 & 86.47 & 99.00 & 96.26 & 87.72 & 84.60 & 80.01 & 74.81\\
    & ss-GAN & 93.12 & 87.20 & 98.93 & 96.70 & 88.40 & 85.27 & 80.42 & 74.58\\
    \bottomrule
    \end{tabularx}
    \caption{Overall accuracies (OA) computed on several datasets with different data augmentation strategies. Sampling strategy is either a 50/50 spatial split of the image (s) or a uniform random sampling of 3\% of the labeled samples (r).}
    \label{table:da_results}
\end{table*}

\section{Data augmentation}
\label{sec:augmentation}

Considering that the synthesized hyperspectral samples are both realistic and diverse, we suggest to use the fake spectra to augment pre-existing hyperspectral datasets. We test this idea on several datasets: Indian Pines (aerial, rural), Pavia University (aerial, peri-urban), Pavia Center (aerial, urban) and Botswana (satellite, rural). Results in the supervised and semi-supervised settings are illustrated in~\cref{table:da_results}. Augmenting the dataset with fake samples marginally increase the classification accuracy when the GAN is trained only on annotated samples. This is expected as the samples would hardly bring new information compared to the true samples. However, training the GAN in a semi-supervised fashion allows us to augment the dataset with fake samples that come from an approximation of the global distribution, including knowledge of how unlabeled samples look like. It therefore increases the model generalization ability, especially in the case where the training and testing set are disjoint.

It is worth noting that increasing drastically the number of fake samples does not increase further the classification accuracy and even degrades it beyond a certain point. We speculate that the introduction of too many approximative samples hurt the model's classification ability.

\section{Discussion}
\label{sec:discussion}

In this work, we presented a method based on Generative Adversarial Networks to generate an arbitrary large number of hyperspectral samples matching the distribution of any dataset, annotated or not. Through a data-driven analysis, we showed that the obtained spectra are plausible as they respect the statistical properties of the real samples. By interpolating between vectors in the latent space, we show that it is possible to synthesize any arbitrary combination of classes, i.e. to perform realistic spectral mixing. This is especially interesting as this could form the basis of data-driven unmixing techniques, e.g. by using a dictionary of synthetic spectra. Finally, we showed that incorporating synthetic samples can serve as a data augmentation strategy for hyperspectral datasets, with positive accuracy improvements on the Indian Pines, Pavia University, Pavia Center and Botswana datasets.

This opens the door to new possibilities in hyperspectral data synthesis and manipulation based on generative models, e.g. domain adaptation by learning the transfer function between two sensors, unmixing by disentangling spectra in the latent domain or hyperspectral data augmentation for deep learning purposes.

\bibliographystyle{IEEEbib}
\bibliography{IGARSS}

\end{document}